\documentclass[10pt,twocolumn]{article}
\usepackage[T1]{fontenc}
\usepackage[utf8]{inputenc}
\usepackage{lmodern}
\usepackage{mathptmx}
\usepackage[letterpaper,margin=0.75in]{geometry}
\usepackage{amsmath,amssymb,mathtools}
\usepackage{graphicx,booktabs,multirow,pifont}
\usepackage[numbers,sort&compress]{natbib}
\usepackage{xurl}
\usepackage[hidelinks]{hyperref}
\usepackage{microtype}
\graphicspath{{figures/}}
\newcommand{\Description}[1]{}

\hypersetup{pdftitle={HGWM: Hierarchical Graph-Guided World Model for Zero-Shot Object Navigation via Graph Matching},pdfauthor={I-Tak Ieong and Hao Tang}}
\begin{document}
\twocolumn[{
\begin{center}
{\LARGE\bfseries HGWM: Hierarchical Graph-Guided World Model for Zero-Shot Object Navigation via Graph Matching\par}
\vspace{1em}
\begin{tabular}{c@{\hspace{3em}}c}
{\large I-Tak Ieong} & {\large Hao Tang\textsuperscript{*}}\\[0.35em]
Tsinghua University & Peking University
\end{tabular}\par
\vspace{0.5em}
{\small \textsuperscript{*}Corresponding author.}\par

\vspace{0.5em}
{\small \href{https://itak04.github.io/HGWM/}{Project page}\quad
\href{https://github.com/itak04/HGWM}{Code}}
\end{center}
\begin{abstract}
Zero-shot object navigation requires an agent to reason about where a target may be found while retaining evidence from previous observations. We present HGWM, a hierarchical graph-guided world model that combines a category-conditioned room--anchor prior with persistent spatial-semantic memory. A coarse stage uses a vision-language model to score candidate headings, while a fine stage explicitly compares observed objects and relations with the goal prior. Phase-dependent fusion integrates these signals with a language-model recommendation to guide local navigation. HGWM achieves 62.6\% success rate (SR) and 31.5\% success weighted by path length (SPL) on HM3D v0.1, improving over WMNav by 4.5 and 0.3 percentage points, respectively. On MP3D, it achieves 46.8\% SR and 17.9\% SPL, with gains of 1.4 and 0.7 points. On HM3D v0.2, the full model reaches 74.4\% SR and 31.9\% SPL; removing individual components reduces SR by 0.9--2.3 points and SPL by 0.6--1.5 points. These results support the use of relational priors and persistent graph memory for target-directed search.
\end{abstract}
\vspace{1em}
}]
\section{Introduction}

Navigation in an unfamiliar environment benefits from both recognizing visible objects and anticipating where a target might be found. Foundation-model agents provide open-vocabulary semantic cues~\citep{yin2025unigoal,nie2025wmnav}, but ambiguous observations and incomplete spatial memory can still lead to unproductive search. A useful prior should guide exploration without being mistaken for a verified map of the environment.

We study two design challenges: retaining observations across viewpoints and using prior room--object relations to assess candidate directions. VLM detections can be incorrect, while common indoor associations need not hold in every scene. Structural checks may therefore help reject some misleading interpretations but can also suppress valid, atypical observations.

HGWM addresses these challenges through a \textit{predict-then-ground} design. An LLM first generates a hierarchical goal graph describing plausible rooms, anchors, and connections. During navigation, RGB-D observations populate a persistent spatial-semantic graph. Coarse visual-semantic scoring and fine prior-conditioned matching then inform action selection. Here, grounding means relating a hypothesis to sensor-derived evidence, not proving that either graph is correct.

A category such as bed specifies no room instance, route, or viewpoint. A room--anchor hypothesis supplies plausible search contexts before the target appears, while memory records encountered evidence. Their compatibility gives relational context a defined role in ranking where to look next.

Our contributions are threefold:
\begin{itemize}
    \item A symbolic predictive navigation state coupling a hierarchical goal prior with persistent spatial-semantic graph memory.
    \item A two-stage decision rule that separates prior-conditioned visual scoring from explicit, direction-specific matching of observed and hypothesized relations.
    \item Evaluation on HM3D and MP3D, including four additional component ablations on HM3D v0.2, analysis of success and path efficiency, and qualitative navigation examples.
\end{itemize}

Relative to WMNav, HGWM improves SR by 4.5 percentage points (pp) on HM3D v0.1 and 1.4 pp on MP3D, with SPL gains of 0.3 and 0.7 pp. We examine the contribution of each component through ablations and use navigation traces to illustrate how the agent responds to changing observations.

\begin{figure*}[t]
\centering
\includegraphics[width=\textwidth]{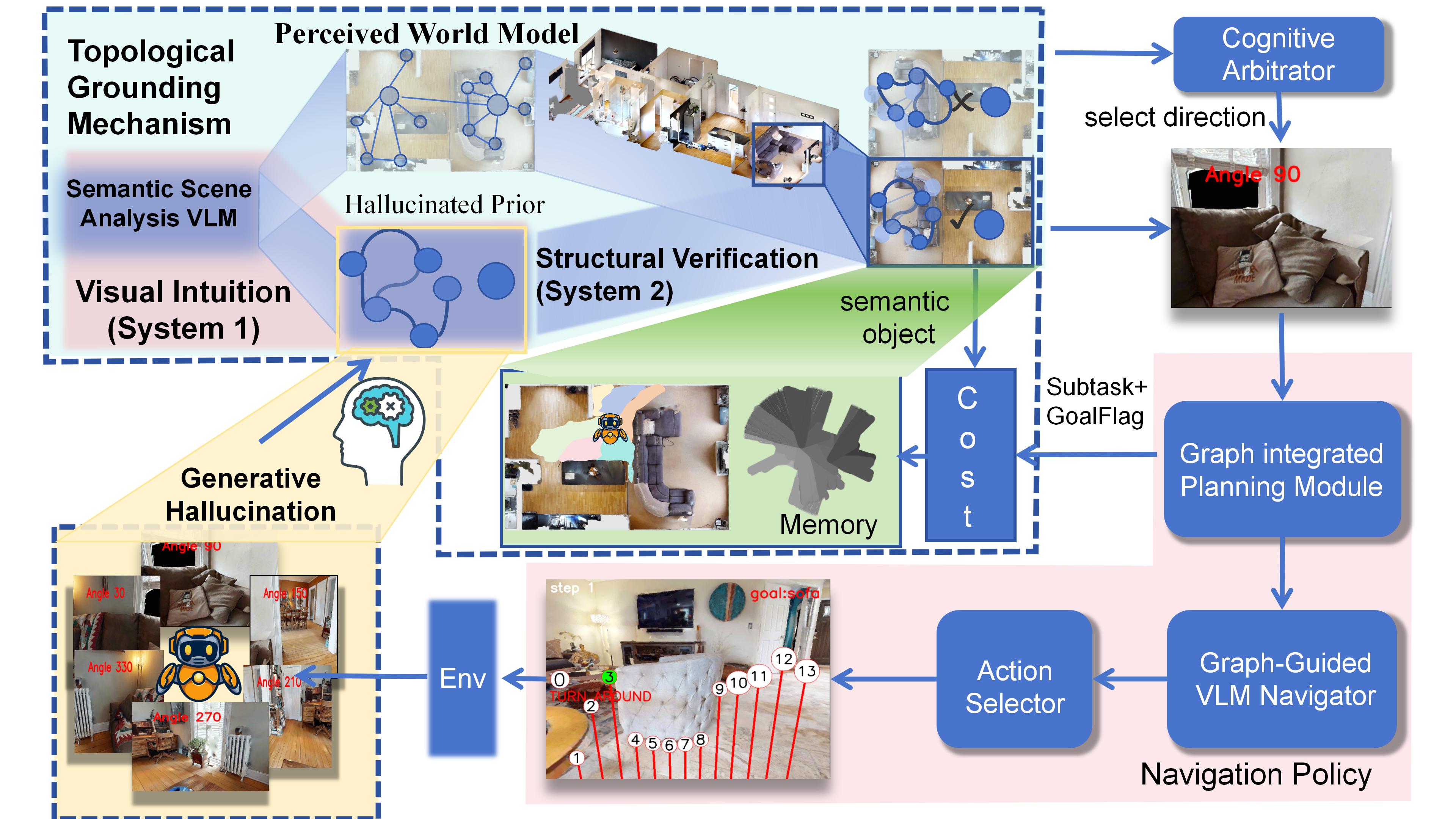}
\caption{HGWM architecture. The LLM generates a hypothetical room--anchor prior, observations populate spatial-semantic memory, and graph-guided reasoning informs navigation. Hallucinated Prior and Structural Verification denote prior generation and heuristic compatibility checks; neither establishes scene truth. System 1/2 labels are conceptual analogies.}
\label{fig:architecture}
\Description{Architecture of HGWM. An LLM produces a predictive goal graph, observations populate persistent spatial-semantic memory, and coarse-to-fine matching and arbitration select navigation actions.}
\end{figure*}

\section{Related Work}

\paragraph{Semantic search and spatial maps.}
ObjectNav methods differ in how they represent prior knowledge and retain observations~\citep{sun2024survey,ieong2025multimodal}. Habitat-Web and OVRL-V2 learn navigation capabilities from task data~\citep{ramrakhya2022habitat,yadav2023ovrl}; ZSON, CoW, ESC, OpenFMNav, and semantic-frontier approaches exploit pretrained representations or commonsense knowledge~\citep{majumdar2022zson,gadre2023cows,zhou2023esc,kuang2024openfmnav,chen2023not}. VLFM accumulates visual-language relevance in a frontier value map~\citep{yokoyama2024vlfm}. VoroNav uses a reduced Voronoi graph and path descriptions for waypoint reasoning~\citep{wu2024voronav}, while BeliefMapNav estimates target-location belief in a 3D voxel space and plans sequences of destinations~\citep{zhou2025beliefmapnav}. Together with 3D-Aware and ApexNav~\citep{zhang20233d,zhang2025apexnav}, these systems establish spatial memory and target-directed search as important design choices. HGWM focuses on a complementary question: how a category-conditioned relational hypothesis should influence the score assigned to each candidate heading.

\paragraph{Scene graphs for embodied reasoning.}
SOON, HOZ, and ConceptGraphs use structured scene representations~\citep{zhu2021soon,zhang2021hierarchical,gu2024conceptgraphs}. SG-Nav builds an online hierarchy of objects, groups, and rooms, uses hierarchical language-model reasoning, and introduces graph-based re-perception~\citep{yin2024sg}. HGWM therefore builds on an established use of graphs for context and perception review. Its specific design couples observed memory with a separate room--anchor hypothesis and makes their compatibility an explicit decision signal alongside visual-semantic scores. The contribution concerns this prior-to-action interface, rather than scene-graph construction alone.

\paragraph{Goal graphs and the ObjectNav setting.}
UniGoal is the closest graph-matching reference~\citep{yin2025unigoal}. It unifies object, image, and text goals, matches node and edge embeddings, evaluates graph-edit similarity, and switches exploration strategies according to the matching state. Its paper explicitly notes that an ObjectNav category becomes a single goal node, disabling goal-graph decomposition and anchor-pair alignment in that setting. HGWM expands the category into room hypotheses, anchor objects, and relations before observation. This expansion supplies relational search cues even when the requested goal is only an object category. HGWM then scores headings with explicit symbolic features and phase-dependent fusion; UniGoal uses matched graph structure and, where applicable, anchor alignment to generate exploration destinations. Both use graph matching and staged reasoning, but they differ in the category-goal representation and the mechanism connecting matching to navigation.

\paragraph{VLM decision memory.}
WMNav scores panoramic views, accumulates a curiosity value map, and combines subtask feedback with a two-stage action proposer~\citep{nie2025wmnav}. HGWM retains this useful map-and-policy structure while adding a separate goal prior, persistent relational evidence, and a fine compatibility channel. Its decision rule combines visual promise, graph compatibility, and an LLM direction recommendation using coefficients selected from the current matching state. This specifies how graph context changes the input to the stored direction-value map; it does not treat memory or subtask decomposition as new in themselves.

\paragraph{Predictive models and HGWM's contribution.}
SGM predicts missing regions of semantic maps~\citep{zhang2024imagine}. World Models, Dreamer, Pathdreamer, and Navigation World Models learn predictive representations of environments~\citep{ha2018world,hafner2023mastering,koh2021pathdreamer,bar2025navigation}. HGWM uses pretrained models to form a symbolic hypothesis and compare it with observed evidence, without training a dynamics model or synthesizing future observations. Predictive coding is a conceptual motivation~\citep{friston2010free,clark2013whatever}. The methodological contribution is the combination of category-to-graph expansion, direction-specific compatibility, and phase-dependent fusion within persistent decision memory. The experiments distinguish whole-configuration performance from evidence about each component.

\section{The Proposed Method}
HGWM maintains a symbolic predictive navigation state: a goal prior, an observation memory, and scores for candidate headings. The term \textit{world model} denotes this state, not a learned dynamics simulator. Figure~\ref{fig:architecture} shows the perception--memory--decision loop. Section~\ref{sec:reproduction} describes the experimental setup.

\subsection{Task and Decision State}
The target is an object category $c$. At decision step $t$, the agent receives RGB-D observations $O_t=(I_t,D_t)$ and a planar pose $P_t\in SE(2)$. The panoramic scorer uses six relative headings $\mathcal A=\{30,90,150,210,270,330\}^{\circ}$. A high-level decision selects a search direction and a short-horizon subtask; the local navigator then selects a feasible displacement. The 40-step horizon counts agent decision iterations, not a uniform metric distance traveled.
\begin{equation}
\mathcal W_t=(G_{goal},M_t,\{(c_t^\alpha,f_t^\alpha)\}_{\alpha\in\mathcal A}).
\end{equation}
Here $c_t^\alpha\in[0,10]$ is a coarse score and $f_t^\alpha\in[0,1]$ is a symbolic compatibility score. Neither is a posterior probability. Success requires the stopping condition within $1.0\,\mathrm m$ of a target instance under the evaluation protocol, rather than complete alignment of the graphs.

\begin{figure*}[t]
\centering
\includegraphics[width=\linewidth]{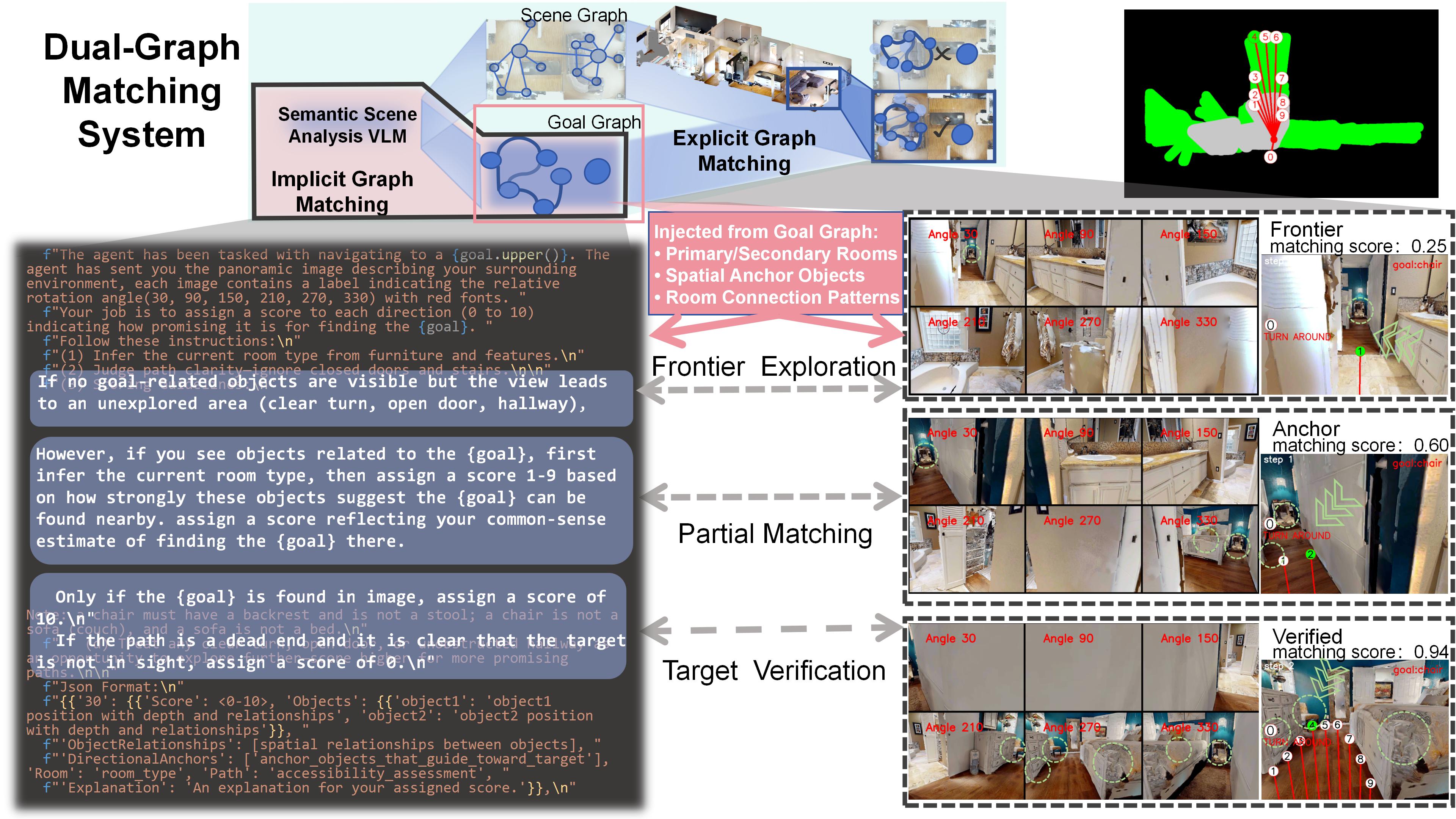}
\caption{Coarse-to-fine illustration with example trajectory views. Coarse VLM scores use 0--10: 0 for a dead end without a visible target, 1--9 for intermediate opportunities, and 10 for a model-reported visible target. The example fine scores are heuristic compatibility values, not calibrated probabilities. The ordered phase thresholds are defined in the text.}
\label{fig:coarse_to_fine}
\Description{Coarse visual-semantic scores and fine symbolic graph-compatibility scores guide exploration, focused search, and target verification.}
\end{figure*}

\subsection{Goal Prior and Observation Memory}
\label{sec:goal_hallucination}
Given $c$, an LLM returns a structured prior containing likely rooms, alternative rooms, anchor objects, and room connections:
\begin{equation}
G_{goal}=\mathrm{LLM}_{prior}(c,\text{prompt}).
\end{equation}
This graph is a category-conditioned hypothesis, not a map of the particular scene. We cache each goal prior within an episode and clear the cache on reset; independent episodes can therefore use different priors.

\paragraph{What is predicted.}
The prior describes object categories and plausible relations, rather than asserting coordinates for unseen objects. For example, a bed query can suggest a bedroom and nearby furniture as search context. A detected anchor contributes evidence for a direction even before the bed is visible; failure to observe that anchor is not proof that a bed is absent. Alternative rooms in the prior preserve multiple search hypotheses. Hypothetical nodes are kept separate from observation records so that an expected object is not counted as a detected object merely because the LLM mentioned it.

\label{sec:world_model}
Spatial-semantic memory stores object names, inferred rooms, observation counts, and estimated map positions. The agent queues depth-based navigation rays before the VLM response arrives, then associates each directional description with the nearest available view-center ray within $30^\circ$. Projection uses pose, a maximum ray distance of $3\,\mathrm m$, and $100$ map pixels per meter. These are coarse positional estimates for a view, not precise object bounding-box projections. The current local graph additionally uses textual spatial and depth cues.

For each direction, current observations and retrieved memory are normalized into one graph contract before matching. It exposes consistent flat and typed views of room nodes, object nodes, adjacency, object relations, and containment. Field aliases are resolved once, objects retain stable identifiers, and observed room membership supplies containment edges. Current attributes take precedence for the same identifier. This boundary prevents a scorer from silently reading a different graph representation. Matching uses symbolic features, without a learned graph-embedding network.

\subsection{Coarse Visual-Semantic Scoring}
\label{sec:grounding}
The VLM receives the panorama, goal, and prior context and returns a JSON record per heading with a score, objects, room, path assessment, and explanation:
\begin{equation}
c_t^\alpha=\mathrm{VLM}_{score}(I_t^{pano},c,G_{goal})_\alpha.
\end{equation}
The requested score range is 0--10. A score of 10 is reserved for a visible target according to the model; 0 denotes a dead end with no visible target; 1--9 encodes intermediate opportunities. This is a model judgment, not a validated detector confidence. Because the same goal prior already enters this prompt, the fine stage adds an explicit computation rather than exclusive access to prior knowledge. The parser supplies 5 when a score is absent, and uses score 5 for all headings if the entire directional extraction fails. The fusion routine consumes this scale directly: it does not apply candidate-set min--max normalization or episode-level z-score normalization. Malformed or out-of-range responses remain an implementation concern, not an assumed probabilistic guarantee.

\subsection{Fine Symbolic Compatibility}
\begin{figure}[t]
\centering
\includegraphics[width=\linewidth]{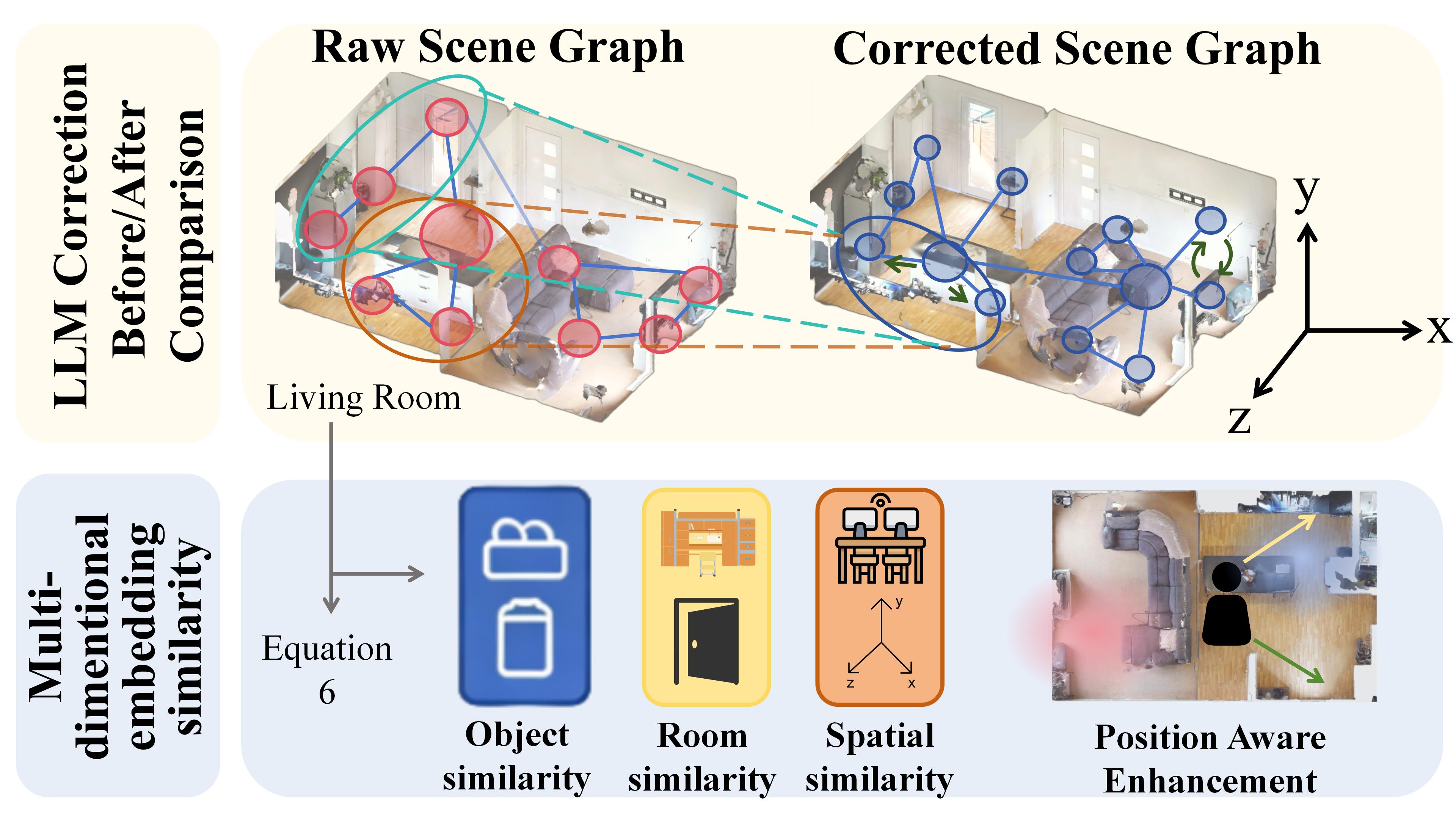}
\caption{Fine-matching illustration. Corrected Scene Graph denotes the graph after heuristic relation proposals, not verified truth. The illustrated embedding similarity is interpreted as symbolic feature comparison, without a learned graph-embedding network. Equation~\ref{eq:fine_matching} specifies all five features, including semantic and depth terms in addition to the illustrated object, room, and spatial terms.}
\label{fig:graph_matching_pipeline}
\Description{Heuristic relation refinement precedes five-component symbolic compatibility scoring and direction conditioning.}
\end{figure}

The refinement prompt asks the LLM to review relations using scene descriptions and goal context. The application routine retains nodes, accepts proposed relations between known nodes, and preserves unmodified edges with confidence greater than 0.7. All typed edge views are rebuilt from the edited edge set, preventing discarded relations from surviving in a stale representation. This operation remains heuristic and can mishandle atypical arrangements; it is not a verified node-pruning procedure.

Let $\mathbf s^\alpha=(s_r,s_o,s_p,s_m,s_d)$ denote five features: room, object, spatial, semantic, and depth. The code compares names, sets, and relations. The room feature combines Jaccard overlap of room names (0.5), overlap of spatial regions (0.3), and connectivity (0.2). The object feature compares visible objects and stored objects with the prior, and with each other. It also rewards matching anchors and rooms. Spatial compatibility combines room relations (0.4), object relations (0.4), and relations between levels (0.2). Each term compares canonical endpoint names and relation labels, dividing the intersection size by the larger relation-set size; absent evidence gives zero. Semantic consistency combines agreement of visible and stored objects (0.4), graph richness (0.3), and goal relevance (0.3). The supplement defines the bonuses, defaults, and weight cases. For name and region sets, the overlap convention is
\begin{equation}
\label{eq:set_overlap}
J(A,B)=\frac{|A\cap B|}{\max\{1,|A\cup B|\}}.
\end{equation}

For depth-layer proportions $\mathbf p$ and reference proportions $\mathbf q=(0.3,0.4,0.3)$, define
\begin{equation}
\label{eq:depth_score}
s_d=\frac13\sum_{k=1}^{3}(1-|p_k-q_k|).
\end{equation}
Missing depth observations give $s_d=0$. The fine score is
\begin{equation}
\label{eq:fine_matching}
f_t^\alpha=\min\{1,(\mathbf w^\top\mathbf s^\alpha+b^\alpha)d^\alpha h^\alpha\}.
\end{equation}
Weights depend on detected prior-associated objects, scene quality, and room/object match scores. The default vector in room, object, spatial, semantic, and depth order is $(0.35,0.30,0.20,0.10,0.05)$. The bonus $b^\alpha$ rewards partial matches and reliable relations. $d^\alpha$ is direction relevance of nearby stored observations, and $h^\alpha$ is a depth-direction factor; both default to 1 when the corresponding memory is unavailable. An exception in overlap calculation returns $f_t^\alpha=0.5$ and records its cause. Empty observations receive no positive graph-match evidence. Per-direction diagnostics record graph sizes, feature scores, and fallback status; the score remains a heuristic rather than a calibrated probability.

\subsection{Phase-Dependent Fusion}
Let $\bar f_t$ and $f_t^{max}$ be the mean and maximum fine scores across the six headings. Target verification takes precedence when $f_t^{max}\geq0.7$; otherwise focused search applies when $\bar f_t\geq0.4$ or $f_t^{max}\geq0.5$; frontier exploration applies otherwise. The LLM receives the directional scene descriptions, scores, and phase, and returns a recommended direction $\hat\alpha_t$ and a scalar $u_t$. The scalar is a reasoning output, not a calibrated confidence interval. Its contribution is
\begin{equation}
\ell_t^\alpha=\begin{cases}u_t,&\alpha=\hat\alpha_t,\\\max(0.2,0.5u_t),&\text{otherwise}.\end{cases}
\end{equation}
For $c_t^\alpha<10$, the fusion rule is
\begin{equation}
\label{eq:fusion}
q_t^\alpha=\mathrm{round}_2\!\left(\mathrm{clip}_{[0,10]}(a c_t^\alpha+10b f_t^\alpha+10e\ell_t^\alpha)\right).
\end{equation}
If $c_t^\alpha\geq10$, the implementation sets $q_t^\alpha=10$. Table~\ref{tab:phase_weights} gives the actual fusion coefficients. They are heuristic coefficients, not convex-mixture probabilities; clipping is therefore part of the definition.

\paragraph{Prior information versus explicit matching.}
Prompting can exploit relations implicitly. HGWM additionally compares named endpoints and relation labels, so equal object sets can receive different spatial scores when their relations differ. At a fixed state with $c_t^{\alpha}<10$, hold the prior, graph, coarse scores, phase coefficients, and LLM output constant. Replacing the fine contribution by an equal-coefficient coarse contribution gives the unclipped diagnostic
\[
S_{\rm ref}^{\alpha}=(a+b)c_t^{\alpha}+10e\ell_t^{\alpha},\qquad
S^{\alpha}-S_{\rm ref}^{\alpha}=b(10f_t^{\alpha}-c_t^{\alpha}).
\]
This separates the numerical matching channel from prior access and coefficient mass. It is an algebraic diagnostic, not a measured navigation ablation; phase selection and LLM outputs can themselves depend on matching. The supplement defines controls for these distinct effects.

\begin{table}[t]
\centering
\caption{Phase-dependent fusion coefficients.}
\label{tab:phase_weights}
\begin{tabular}{lrrr}
\toprule
Phase & $a$ (coarse) & $b$ (fine) & $e$ (LLM)\\
\midrule
Frontier exploration & 0.6 & 0.1 & 0.5\\
Focused search & 0.6 & 0.4 & 0.4\\
Target verification & 0.8 & 0.5 & 0.2\\
\bottomrule
\end{tabular}
\end{table}

\subsection{Memory Update and Local Action Selection}
Fused directional values update the spatial curiosity map. Regions shared by adjacent headings receive the mean of their values; the base policy merges them with stored map values and selects the heading with the greatest regional mean. Thus selecting the largest instantaneous $q_t^\alpha$ alone does not fully specify the policy. A planning VLM uses the selected view, previous subtask, goal, and graph context to propose the next subtask and goal flag. The local action selector reasons over annotated feasible displacements; depth-derived navigability constrains motion. Stopping and navigation remain model-driven decisions and can fail despite a high score. The accompanying code provides the prompt builders and scoring routines.

More precisely, let $V_t(x)$ denote the stored curiosity value at map cell $x$, and let $v_t(x)$ be the fused value projected from the current panorama. For cells updated by the base policy, the merge is
\begin{equation}
V_t(x)=\min\{V_{t-1}(x),v_t(x)\}.
\end{equation}
Cells outside the update mask retain their previous values; cells in the goal mask are assigned 10. A valid directional region is ranked by its mean stored value. This inherited conservative update makes the decision depend on previously accumulated evidence, not just the latest panorama. It can also preserve an erroneous low score, which motivates evaluating failures at the trajectory level.

\subsection{Execution and Information Flow}
Each navigation decision follows the same information flow:
\begin{enumerate}
    \item Obtain the panorama, depth, and pose; generate or reuse the goal prior for the current episode.
    \item Request the six visual direction scores together with object, room, and path descriptions.
    \item Associate semantic descriptions with spatial memory and form direction-specific graph evidence. Where enabled, request relation proposals before computing compatibility features.
    \item Compute the fine scores, choose the exploration phase, and obtain the joint LLM direction recommendation. Apply the phase coefficients and score clipping.
    \item Update the curiosity map, select a direction, and use the chosen view and previous subtask to produce the next local objective.
    \item Select a feasible displacement from the annotated action candidates, execute it, and use the next observation to continue or stop.
\end{enumerate}
This ordering separates the goal hypothesis, observed evidence, direction ranking, and motion interface. It also identifies what an ablation changes: the prior affects contextual conditioning; relation refinement affects graph evidence; arbitration affects the fused score; and weight adaptation changes the fine feature mixture. None of these changes requires treating an LLM score as a calibrated probability.

\section{Experiments}

\begin{figure*}[t]
    \centering
    \includegraphics[width=0.98\linewidth]{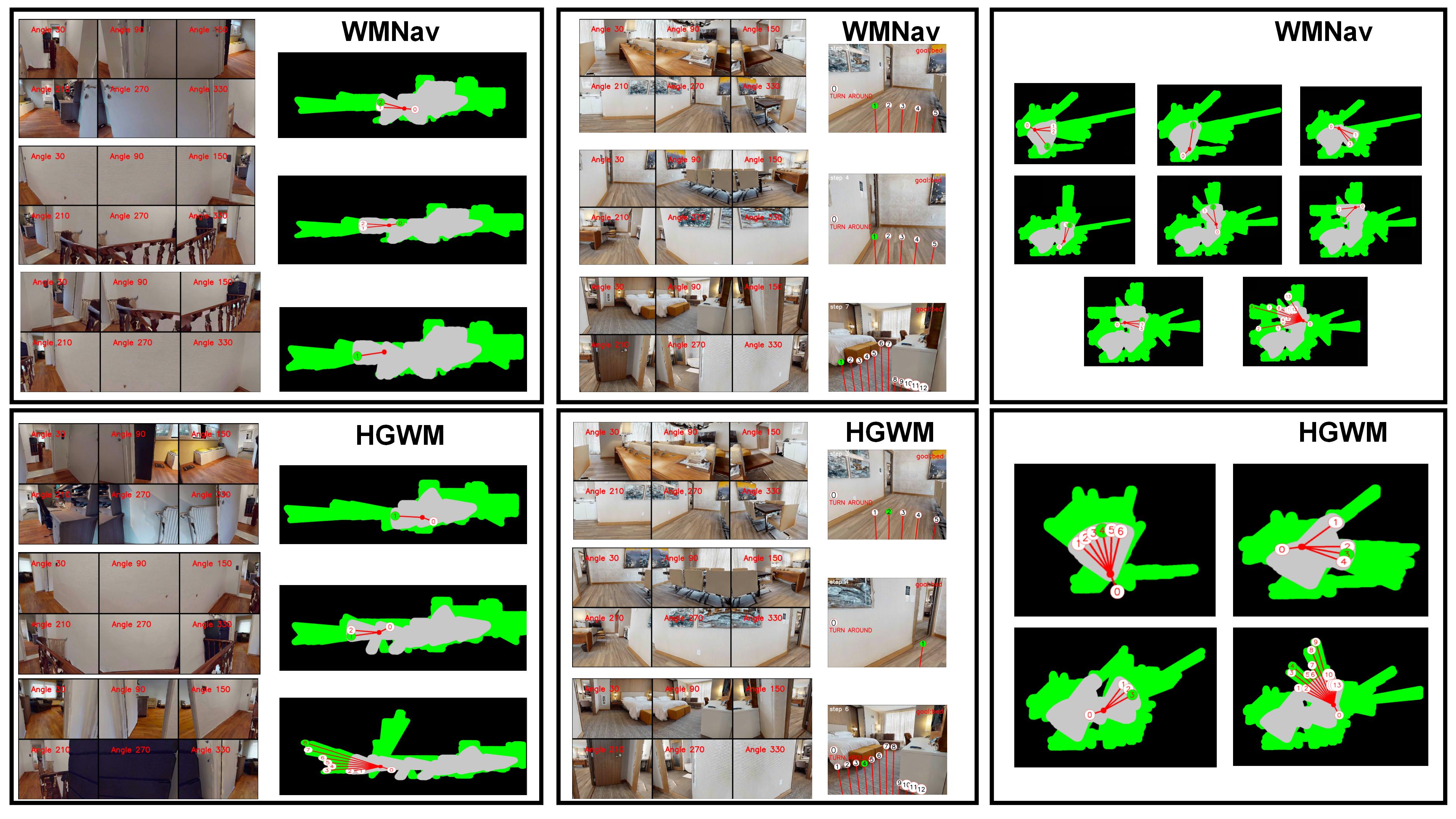}
    \caption{Decision process comparison between WMNav and HGWM. In these selected examples, HGWM visits different regions and exhibits less repeated exploration. These trajectories illustrate behavior; they do not establish an aggregate reduction in looping or isolate the effect of graph matching.}
    \label{fig:comparison}
\Description{Qualitative navigation comparison between WMNav and HGWM, showing repeated exploration by WMNav and graph-memory-guided decisions by HGWM.}
\end{figure*}

We report simulation experiments on HM3D and MP3D to evaluate the complete HGWM configuration. The experiments do not include physical-robot trials or a systematic sensor-noise robustness protocol.

\subsection{Datasets and Evaluation Metrics}

Datasets. We evaluate on HM3D v0.1, HM3D v0.2, and MP3D~\citep{ramakrishnan2021habitat,chang2017matterport3d}. We report main comparisons on HM3D v0.1 for consistency with prior work, and use HM3D v0.2 for the three-configuration component comparison. Results from these splits should not be pooled as if they used the same episodes.

Evaluation metrics. We report Success Rate (SR) and Success weighted by Path Length (SPL), following standard ObjectNav evaluation protocols~\citep{anderson2018evaluation,batra2020objectnav}. These metrics measure navigation effectiveness and efficiency, respectively. Figure~\ref{fig:comparison} illustrates selected trajectories; repeated exploration is not reported as a quantitative metric.

\paragraph{Reading SR and SPL together.}
For $N$ episodes, let $s_i$ indicate success, $l_i$ be the shortest valid path length, and $p_i$ be the traveled path length. The reported percentage metrics are
\begin{equation}
\begin{aligned}
\mathrm{SR}&=\frac{100}{N}\sum_{i=1}^{N}s_i,\\
\mathrm{SPL}&=\frac{100}{N}\sum_{i=1}^{N}s_i\frac{l_i}{\max(l_i,p_i)}.
\end{aligned}
\end{equation}
SR measures completion frequency; SPL also penalizes excess travel on successful episodes. A higher SR therefore need not imply a higher SPL. All differences below are absolute percentage-point changes, computed within a dataset and split; HM3D v0.1 and v0.2 results are kept separate.

\subsection{Implementation and Reproducibility}
\label{sec:reproduction}

The agent is configured for at most 40 navigation steps per episode, with a cylindrical body of radius $0.18\,\mathrm{m}$ and height $0.88\,\mathrm{m}$. Its RGB-D camera has a resolution of $640\times480$, a $79^{\circ}$ horizontal field of view, and a $14^{\circ}$ downward pitch. The success threshold is $d_{thres}=1.0\,\mathrm{m}$. The navigation-map scale is 100 pixels per meter.

Gemini-1.5-Pro provides scene understanding and coarse scores, while Qwen2.5-7B-Instruct generates goal priors, proposes relation updates, and recommends directions. Goal priors are cached within each episode. The implementation records configuration and source identifiers, episode identities, model-call counts, and elapsed time. We use HM3D v0.1 and MP3D for comparisons with prior methods and HM3D v0.2 for component analysis. For the main HGWM setting on HM3D v0.1, we also summarize variation across three runs.

\subsection{Main Results}

\begin{table*}[t]
\caption{Comparison on HM3D v0.1 and MP3D. SR and SPL are percentages; higher is better. Baseline results are taken from the cited papers and use their respective training and model settings. For HGWM on HM3D v0.1, superscripts and subscripts give the maximum and minimum deviations across three runs, not confidence intervals.}
\label{tab:main_results}
\centering
\small
\begin{tabular*}{\textwidth}{@{\extracolsep{\fill}}lllrrrr@{}}
\toprule
\multirow{2}{*}{\textbf{Method}} & \multirow{2}{*}{\textbf{Vision}} & \multirow{2}{*}{\textbf{Language}} & \multicolumn{2}{c}{\textbf{HM3D v0.1}} & \multicolumn{2}{c}{\textbf{MP3D}} \\
& & & \textbf{SR$\uparrow$} & \textbf{SPL$\uparrow$} & \textbf{SR$\uparrow$} & \textbf{SPL$\uparrow$} \\
\midrule
Habitat-Web~\citep{ramrakhya2022habitat} & - & - & 41.5 & 16.0 & 31.6 & 8.5 \\
OVRL~\citep{yadav2022ovrl} & - & - & - & - & 28.6 & 7.4 \\
OVRL-V2~\citep{yadav2023ovrl} & - & - & 64.7 & 28.1 & - & - \\
ZSON~\citep{majumdar2022zson} & CLIP & - & 25.5 & 12.6 & 15.3 & 4.8 \\
PSL~\citep{sun2024prioritized} & CLIP & - & 42.4 & 19.2 & 18.9 & 6.4 \\
PixNav~\citep{cai2024bridging} & LLaMA-Adapter & GPT-4 & 37.9 & 20.5 & - & - \\
SGM~\citep{zhang2024imagine} & - & - & 60.2 & 30.8 & 37.7 & 14.7 \\
VLFM~\citep{yokoyama2024vlfm} & BLIP-2 & - & 52.5 & 30.4 & 36.4 & 17.5 \\
\addlinespace
CoW~\citep{gadre2023cows} & CLIP & - & - & - & 9.2 & 4.9 \\
ESC~\citep{zhou2023esc} & - & GPT-3.5 & 39.2 & 22.3 & 28.7 & 14.2 \\
L3MVN~\citep{yu2023l3mvn} & - & GPT-2 & 50.4 & 23.1 & - & - \\
VoroNav~\citep{wu2024voronav} & BLIP & GPT-3.5 & 42.0 & 26.0 & - & - \\
HyPerNav~\citep{yin2025hypernav} & Qwen-VL-7B-Instruct & - & 56.3 & 29.7 & 43.8 & 16.9 \\
TopV-Nav~\citep{zhong2024topv} & - & - & 45.9 & 28.0 & 31.9 & 16.1 \\
SG-Nav~\citep{yin2024sg} & LLaVA-1.6-7B & GPT-4 & 54.0 & 24.9 & 40.2 & 16.0 \\
UniGoal~\citep{yin2025unigoal} & LLaVA-1.6-7B & LLaMA-2-7B & 54.5 & 25.1 & 41.0 & 16.4 \\
WMNav~\citep{nie2025wmnav} & Gemini-1.5-Pro & - & 58.1 & 31.2 & 45.4 & 17.2 \\
\addlinespace
\textbf{HGWM (Ours)} & \textbf{Gemini-1.5-Pro} & \textbf{Qwen2.5-7B-Instruct} & $\mathbf{62.6}^{+0.2}_{-0.3}$ & $\mathbf{31.5}^{+0.2}_{-0.3}$ & \textbf{46.8} & \textbf{17.9} \\
\bottomrule
\end{tabular*}
\end{table*}

Table~\ref{tab:main_results} reports HGWM at 62.6\% SR and 31.5\% SPL on HM3D v0.1, compared with WMNav's 58.1\% and 31.2\%: gains of 4.5 pp SR and 0.3 pp SPL. On MP3D, HGWM reports 46.8\% SR and 17.9\% SPL versus 45.4\% and 17.2\%, giving 1.4 pp SR and 0.7 pp SPL. In particular, the path-efficiency gains are modest. OVRL-V2 has a higher reported HM3D SR (64.7\%) but uses a different training regime.

Across three HGWM runs on HM3D v0.1, SR ranges from 62.3 to 62.8 and SPL from 31.2 to 31.7. These observed ranges are not confidence intervals. The SPL range reaches WMNav's point estimate of 31.2, so the clearest improvement is in completion rate rather than a consistent gain in path efficiency.

\begin{table}[t]
\centering
\caption{Differences between configurations in percentage points, computed from the corresponding SR and SPL values.}
\label{tab:reported_deltas}
\begin{tabular}{lrr}
\toprule
Comparison & $\Delta$SR & $\Delta$SPL\\
\midrule
HGWM--WMNav, HM3D v0.1 & +4.5 & +0.3\\
HGWM--WMNav, MP3D & +1.4 & +0.7\\
Voxel--no memory, HM3D v0.2 & +3.1 & +0.8\\
HGWM--voxel, HM3D v0.2 & +3.5 & +1.8\\
\bottomrule
\end{tabular}
\end{table}
Table~\ref{tab:reported_deltas} summarizes the gains in success rate and SPL within each dataset.

\subsection{Qualitative Navigation Behavior}

Figure~\ref{fig:comparison} illustrates selected WMNav and HGWM trajectories. The supplementary case collection contains 18 navigation traces, covering 338 step directories and 337 text records. Each case includes its available step logs and views selected by the same first/middle/last rule. These examples show the observations, model responses, and decisions underlying a search trajectory; they complement the quantitative evaluation rather than estimate loop frequency.

\subsection{Backbone Sensitivity}

Table~\ref{tab:vlm_robustness} compares two VLM settings on HM3D v0.1. Replacing Gemini-1.5-Pro with Gemini 2.0 Flash changes SR from 62.6\% to 59.2\% and SPL from 31.5\% to 30.8\%, decreases of 3.4 and 0.7 pp. The larger change in SR suggests that successful search is more sensitive to the backbone in this comparison. Evaluating additional model families would help determine how broadly this pattern holds.

\begin{table}[t]
\centering
\caption{Two-backbone sensitivity on HM3D v0.1. Changes from the default are in percentage points (pp).}
\label{tab:vlm_robustness}
\begin{tabular}{lrr}
\toprule
\textbf{VLM Backbone} & \textbf{SR (\%)} & \textbf{SPL (\%)} \\
\midrule
Gemini-1.5-Pro & \textbf{62.6} & \textbf{31.5} \\
Gemini 2.0 Flash & 59.2 & 30.8 \\
Change (pp) & $-3.4$ & $-0.7$ \\
\bottomrule
\end{tabular}
\end{table}

\subsection{Component Analysis on HM3D v0.2}
Table~\ref{tab:ablation_study} compares three memory configurations and four component ablations on HM3D v0.2. The full model achieves 74.4\% SR and 31.9\% SPL. Basic VLM navigation obtains 67.8/29.3, while voxel memory with direct VLM scoring obtains 70.9/30.1. The configuration comparison changes both memory representation and decision scoring; the component ablations examine specific parts of the full system.

\begin{table*}[t]
\centering
\caption{Configurations and component ablations on HM3D v0.2. Metrics are percentages; differences are percentage points relative to full HGWM. The first two baselines change the overall configuration. Non-adaptive weights denote a single control, not separate uniform and static experiments.}
\label{tab:ablation_study}
\begin{tabular*}{\textwidth}{@{\extracolsep{\fill}}lrrrr@{}}
\toprule
Configuration & SR (\%) & SPL (\%) & $\Delta$SR & $\Delta$SPL\\
\midrule
Basic VLM / no memory & 67.8 & 29.3 & $-6.6$ & $-2.6$\\
Voxel memory + direct VLM scoring & 70.9 & 30.1 & $-3.5$ & $-1.8$\\
\textbf{HGWM (full)} & \textbf{74.4} & \textbf{31.9} & 0.0 & 0.0\\
\addlinespace
Without goal prior & 72.1 & 30.8 & $-2.3$ & $-1.1$\\
Without relation refinement & 72.1 & 30.4 & $-2.3$ & $-1.5$\\
Without LLM arbitration & 72.7 & 31.1 & $-1.7$ & $-0.8$\\
With non-adaptive fine weights & 73.5 & 31.3 & $-0.9$ & $-0.6$\\
\bottomrule
\end{tabular*}
\end{table*}

\paragraph{Prior and relation evidence.}
Removing the goal prior yields 72.1\% SR and 30.8\% SPL, decreases of 2.3 and 1.1 points. Disabling relation refinement gives 72.1\% SR and 30.4\% SPL, decreases of 2.3 and 1.5 points. The larger SPL drop supports the contribution of relation processing at the aggregate level. The prior supplies scoring context; refinement changes matching evidence. Paired trajectories are needed to determine which episodes benefit.

\paragraph{Meaning of the interventions.}
The no-prior control retains the target category but uses an unknown room, empty room connections, and a minimal goal anchor. It removes the explicit room--anchor hypothesis while retaining the pretrained models' background knowledge. The no-refinement control bypasses relation updates. The no-arbitration control removes the LLM term without renormalizing the remaining coefficients, so it changes both information and score scale. The non-adaptive control changes the five fine-feature weights rather than the phase-dependent fusion coefficients.

\paragraph{Arbitration and adaptive weighting.}
Without LLM arbitration, performance is 72.7/31.1, below the full model by 1.7 SR and 0.8 SPL points. The non-adaptive fine-weight control gives 73.5/31.3, decreases of 0.9 and 0.6 points. Both results favor the full fusion design in this comparison. Their relative drops should not be interpreted as additive contributions: each intervention is evaluated against the full configuration, and joint removals are not reported. Separate uniform and static controls would be needed to compare those two choices.

\paragraph{Separating information from score scale.}
A scale-matched counterpart to the no-arbitration control can be defined as
\[
(a',b',e')=\frac{a+b+e}{a+b}(a,b,0).
\]
This preserves the total coefficient mass while removing the LLM term. The no-arbitration result in Table~\ref{tab:ablation_study} uses the unnormalized removal. The scale-matched alternative offers a way to separate the loss of reasoning information from the effect of rescaling the fused scores. The supplement discusses analogous controls for graph matching.

\paragraph{Success, efficiency, and attribution.}
Full HGWM improves over voxel memory by 3.5 SR and 1.8 SPL points. In a separate comparison, scene--goal graph matching improves SR by 1.5 points and SPL by 1.4 points over prior prompting. These gains suggest that computing compatibility between observed and expected relations can provide useful information beyond including the prior in a prompt. The latter comparison is summarized by aggregate differences rather than split-specific absolute scores, so it provides supporting evidence rather than a complete isolation of all downstream matching effects.

\subsection{Efficiency and Limitations}

Gemini-1.5-Pro inference takes approximately 2.3 seconds per call, while graph updates take 12--25 ms at the tested episode lengths. Model inference therefore dominates the graph-processing cost. Reusing the goal prior within an episode avoids repeated prior-generation calls. These component timings do not measure complete navigation latency, which also includes perception, planning, communication, and motion.

\paragraph{Query and latency accounting.}
One panoramic response covers six headings; prior generation, relation reasoning, arbitration, and planning can add calls. Thus per-call latency is not full-step latency. The runner records client calls and elapsed time by role. Provider-internal retries, token usage, simulation time, and motion time still require separate accounting; using the same evaluation episodes does not ensure equal inference cost.

\paragraph{Evidence and robustness limits.}
Our experiments use static Habitat environments. Physical deployment introduces localization and depth noise, moving objects, and atypical room layouts that may conflict with the prior. Errors in perception or pose can also persist in memory. The present comparisons do not quantify these effects or provide paired significance tests. The non-adaptive result does not distinguish uniform from static weighting, and the matching gain is summarized without split-specific absolute scores. Broader robustness tests, matched-cost comparisons, and complete per-run summaries would strengthen these findings.

\section{Conclusion}

HGWM expands category goals into relational priors and combines observed graph memory with scoring on a 0--10 decision scale. Simulation results show SR gains over WMNav and smaller SPL gains. Additional HM3D v0.2 interventions reduce SR by 0.9--2.3 points and SPL by 0.6--1.5 points relative to full HGWM. These findings support the combined design; component interactions, paired uncertainty, query cost, and deployment beyond simulation remain open evaluation questions.

\paragraph{Code Availability.}
Code, prompt builders, and the case gallery are available at \url{https://github.com/itak04/HGWM} and \url{https://itak04.github.io/HGWM/}.

\clearpage
\appendix
\section*{Supplementary Material}
\section{Goal Prior Construction}
\label{sec:goal_construction}

\subsection{LLM Prompt Template}

We present the high-level strategy template for goal subgraph construction; the source index identifies the executable builder, including its output schema and examples. The template is dynamically instantiated for each target object category, with \texttt{\{goal\}} replaced by the specific target (e.g., ``bed'', ``sofa'').

\begin{quote}
You are an expert in residential navigation and spatial understanding. Create a comprehensive navigation strategy for efficiently locating a \{GOAL\} in a typical home environment.

NAVIGATION STRATEGY ANALYSIS FOR \{GOAL\}:

1. TARGET ROOM ANALYSIS:\\
~~(1) Primary room: The most likely room type where \{goal\} is typically found\\
~~(2) Secondary rooms: 1-2 alternative room types where \{goal\} might also be located\\
~~(3) Adjacent zones: Connected areas that provide visual access or pathways to target rooms

2. PATHWAY and CONNECTIVITY MAPPING:\\
~~(1) Hallway connections: Common routes to reach target rooms from main areas\\
~~(2) Visual access points: Doorways/openings that help identify target rooms\\
~~(3) Navigation landmarks: Key reference points for wayfinding

3. TARGET-ORIENTED SPATIAL ANCHORS\\and NAVIGATION CUES:\\
~~(1) Directional anchors: 3-4 objects that DIRECTLY indicate where \{goal\} is likely located\\
~~(2) Visual pathways: Objects that form visual lines leading toward typical \{goal\} placement areas\\
~~(3) Distance indicators: Objects that help estimate proximity to \{goal\}\\
~~(4) Positioning cues: Furniture/fixtures indicating the specific area where \{goal\} is usually found

4. SYSTEMATIC EXPLORATION APPROACH:\\
~~(1) Exploration priorities: Which areas to check first when searching\\
~~(2) Recovery strategies: What to do when initial search paths fail
\end{quote}

This structured prompt enables the LLM to generate hierarchical spatial knowledge encompassing room classifications, connectivity patterns, and object-based navigation cues, which are then parsed into the formal goal subgraph representation $G_{goal} = (V_{goal}, E_{goal})$ where nodes represent rooms/objects and edges encode spatial relationships.

\section{Implementation Details}
\label{app:implementation}

\subsection{Agent Configuration}
We configure the agent with a maximum of 40 navigation steps per episode. The agent has a cylindrical body with radius 0.18m and height 0.88m, equipped with an egocentric RGB-D camera at 640$\times$480 resolution and 79$^{\circ}$ horizontal FOV. The camera is tilted down with a 14$^{\circ}$ pitch to improve navigable area identification. The success threshold $d_{thres}$ is set to 1.0m. Gemini-1.5-Pro is the primary VLM used for the reported experiments.

\subsection{Model Choices}
\textbf{VLM Backbone:} Gemini-1.5-Pro serves as our primary vision-language model for both scene understanding (object detection, room classification) and coarse semantic scoring. The reported experiments do not include a matched cost or spatial-reasoning comparison against GPT-4V.

\textbf{LLM Reasoner:} Qwen2.5-7B-Instruct generates goal priors, proposes relation updates, and recommends navigation directions. The implementation accesses the model through the SiliconFlow API. Its calls contribute to the overall reasoning cost alongside VLM perception and planning.

\section{Coarse-to-Fine Reasoning}
\label{app:multistage}

\subsection{Score Domains and Phase Selection}
For the six panorama headings, the coarse VLM prompt requests scores in $[0,10]$. A model-reported visible target receives 10; an unpromising dead end without a visible target receives 0; intermediate opportunities receive 1--9. Missing individual scores default to 5, and failed directional extraction defaults all six headings to 5. A target score is not a ground-truth detection.

The fine overlap is $f^\alpha\in[0,1]$ in the intended score path. Let $\bar f$ and $f^{max}$ denote the mean and maximum over headings. The implemented schedule is ordered as follows:
\begin{enumerate}
\item If $f^{max}\geq0.7$, use target verification.
\item Otherwise, if $\bar f\geq0.4$ or $f^{max}\geq0.5$, use focused search.
\item Otherwise, use frontier exploration.
\end{enumerate}
The main paper lists the three coefficient triples for verification, focused search, and exploration. The source names them spatial, semantic, and LLM weights. Here the first coefficient multiplies the coarse VLM score, despite its name; it does not multiply a separate depth score.

The arbitrator returns a recommended heading $\hat\alpha$ and a scalar $u$. The boost is $u$ for the recommended heading and $\max(0.2,u/2)$ elsewhere. Fusion multiplies the fine score and LLM boost by 10, adds the weighted coarse score, clips to $[0,10]$, and rounds to two decimal places. A coarse score at least 10 overrides this rule with 10. The coefficients do not sum to one. No candidate-set or episode-level normalization, estimated confidence bounds, or additional position-score multiplier is applied by this routine. Equal directional scores therefore require no division by a zero range or variance. Final heading selection also depends on the stored curiosity map.

\subsection{Symbolic Feature Definitions}
All name-based set comparisons below lowercase strings. Define $J(A,B)=|A\cap B|/\max(1,|A\cup B|)$, so two empty sets have overlap zero. These are symbolic matching operations, not cosine similarity of learned embeddings.

\paragraph{Room feature $s_r$.}
The prior room set contains names in its \texttt{nodes} list with one of the substrings kitchen, bathroom, bedroom, living, dining, or office. The observed set uses \path{room_nodes}. Known spatial-region labels form a second pair of sets. The score is $0.5J(\text{rooms})+0.3J(\text{regions})+0.2C$, capped at one, where $C$ is the fraction of room pairs passing the room-connection predicate. For fewer than two observed rooms, we set $C=0$. The remaining connectivity predicate is heuristic, not a verified floor-plan relation.

\paragraph{Object feature $s_o$.}
Let $V$ be current VLM object names, $H$ observed graph object names, and $T$ the union of the requested category, prior anchors, and named prior object nodes. Let $P\subseteq H$ contain objects in matched rooms; if no rooms match, all observed graph objects are prioritized. Put $n=|V\cap T|$ and
\[
\begin{aligned}v=\min\{1,\;&J(V,T)+\min(0.6,0.5n)\\&+0.2\mathbb I[T\ne\varnothing,\ T\subseteq V]\}.\end{aligned}
\]
with $v=0$ when $n=0$. The stored-object score is
\[
h=\min\{1,J(H,T)+0.2|P\cap T|/\max(1,|T|)\}.
\]
The object feature is $\min\{1,0.7v+0.2h+0.1J(V,H)+0.1s_r\}$ if $n>0$, and $\min\{1,0.4v+0.4h+0.2J(V,H)+0.1s_r\}$ otherwise. The source's \path{target_detected} flag means any member of $T$ matches, including an anchor; it is not exclusively detection of the navigation target.

\paragraph{Spatial feature $s_p$.}
The spatial scorer uses $s_p=0.4r+0.4o+0.2x$. The three terms compare room, object, and containment relation sets respectively. A relation key consists of canonical endpoint names and a relation label, so different instance IDs do not prevent matching. Known symmetric relations (near, adjacent, connected) ignore endpoint order. Each score is the intersection size divided by the larger set size; a missing observed or expected set gives zero.

\paragraph{Semantic and depth features.}
The semantic feature is the capped weighted sum
\[s_m=\min\{1,0.4A+0.3R+0.3G\},\]
where $A$ measures VLM--graph semantic alignment, $R$ measures hierarchical semantic richness, and $G$ measures goal relevance. Specifically, $A=J(V,H)$ with $A=0$ for two empty sets. For room, object, and total edge counts $n_r,n_o,n_e$, richness is
\[
R=\tfrac12\left[\min\{1,(0.3n_r+0.1n_o)/10\}+\min\{1,n_e/20\}\right].
\]
Goal relevance divides the number of matching observed objects by the number of named prior nodes, capped at one. Exact name matches count as 1 and predefined semantic-group matches as 0.5. No named prior nodes gives $G=0.5$. The groups are seating/work furniture; bedding; kitchen fixtures; bathroom fixtures; and books/shelves. The exact group members and predicates are listed in the source index. Exceptions in these three subterms return 0.5. These are deterministic heuristics, not a learned semantic model. For foreground, mid-ground, and background proportions $p_k$, the depth term is $s_d=\frac13\sum_k(1-|p_k-q_k|)$ with $\mathbf q=(0.3,0.4,0.3)$. When no depth-layer observations are present, the depth score is $s_d=0$ rather than a near-reference default distribution.

\subsection{Fine Weights, Bonuses, and Direction Factors}
The first satisfied row of Table~\ref{tab:fine_weights} determines the weight vector. Scene quality is
\[
Q=0.4(0.6v_{frac}+0.4h_{frac})+0.3\bar c_{rel}+0.3\bar c_{room},
\]
capped at one. Here $v_{frac}$ and $h_{frac}$ are the fractions of voxel-validated and high-precision objects, and the two means summarize relation and room confidence fields. Terms with no supporting objects/relations/rooms contribute zero.

\begin{table*}[t]
\centering
\caption{Fine-feature weights used by the fine-scoring routine, in first-match order. These are implementation settings, not fitted performance results.}
\label{tab:fine_weights}
\begin{tabular}{lrrrrr}
\toprule
Condition & Room & Object & Spatial & Semantic & Depth\\
\midrule
Any prior-associated object detected & 0.25 & 0.50 & 0.20 & 0.03 & 0.02\\
Scene quality $Q>0.8$ & 0.35 & 0.30 & 0.25 & 0.05 & 0.05\\
Room score $s_r>0.7$ & 0.45 & 0.25 & 0.20 & 0.05 & 0.05\\
Object score $s_o>0.7$ & 0.30 & 0.40 & 0.20 & 0.05 & 0.05\\
Otherwise & 0.35 & 0.30 & 0.20 & 0.10 & 0.05\\
\bottomrule
\end{tabular}
\end{table*}

The additive bonus has three independently applicable terms: (i) $\min(0.15,0.2s_os_d)$ if $0.3\leq s_r<0.8$, $s_o>0.5$, and $s_d>0.6$; (ii) $\min(0.15,0.15s_r+0.1Q)$ if $0.3\leq s_o<0.8$ and $s_r>0.5$; and (iii) $\min(0.1,0.1\bar c_{rel})$ if relations exist and their mean confidence exceeds 0.7.

For a stored map cell $i$ at distance $r_i<200$ map pixels, let $\delta_i$ be the wrapped angular difference to the candidate heading, $v_i=\max(0.1,1-r_i/200)$, and $n_i=\min(1,N_i/5)$ for observation count $N_i$. Direction relevance is
\[
d^\alpha=\frac{\sum_i\max(0,1-|\delta_i|/\pi)v_in_i}{\max(0.1,\sum_i v_in_i)}.
\]
It defaults to 1 without stored semantic memory or an available agent pose. The depth-direction factor is $h^\alpha=\min(1.2,p_Fu_F+p_Bu_B+1-p_F-p_B)$: $(u_F,u_B)=(1.2,0.8)$ for 30/330 degrees, $(0.9,1.1)$ for 90/270 degrees, and $(1,1)$ otherwise. It defaults to 1 without depth-layer objects. The fine routine receives its direction argument from the caller; the primary scoring caller converts numeric degree strings directly to degree values.

\subsection{Memory, Prompts, and Execution}
The default configuration specifies a navigation map scale of 100 pixels per meter, high-level horizon 40, RGB-D resolution $640\times480$, horizontal field of view $79^\circ$, camera pitch $-0.25$ radians, body radius $0.18\,\mathrm m$, body height $0.88\,\mathrm m$, and success distance $1.0\,\mathrm m$.

Goal priors are generated at runtime, cached within an episode, and cleared on reset. Provider-specific sampling settings are defined in \texttt{src/api.py}. Because independent episodes can use different priors and model outputs, the pipeline does not assume deterministic inference across runs.

The prompt template above describes the goal-prior construction strategy. The executable builders specify the output schemas, examples, and error handling for prior generation, panorama scoring, relation refinement, arbitration, and local planning.

\subsection{Graph Representation and Memory Updates}
Before scoring, the graph interface normalizes node and relation fields and merges retrieved memory with current observations. Flat and typed edge collections represent the same edited graph. Room aliases and goal-schema aliases are resolved consistently, while canonical endpoint names support matching across different instance identifiers. Missing observation, relation, or depth evidence contributes zero to its corresponding match term.

Depth-ray positions are stored until the panoramic VLM response is available. Object descriptions are associated with the closest view-center ray and then added to persistent memory. The resulting positions are coarse view-level estimates. Planning uses the room descriptions and heading angles expressed in degrees.

\subsection{Ablation Settings and Evaluation Tools}
Uniform fine weights assign 0.2 to each feature. Static weights normalize a configured vector; the default vector is
\[
(0.30,0.40,0.20,0.05,0.05),
\]
ordered by room, object, spatial, semantic, and depth features. These are separate modes. The non-adaptive result in Table~\ref{tab:additional_components} represents one configuration and does not distinguish these two modes.

The no-arbitration control removes the LLM contribution without rescaling the other terms. A separate scale-matched variant multiplies the remaining coefficients by $(a+b+e)/(a+b)$. The prior-prompt-only setting bypasses graph scoring and graph context in planning; the graph-memory/direct-scoring setting retains graph context while using raw visual direction scores. These broader settings should be distinguished from an intervention that changes only the numerical matching term.

Runs save manifests with source hashes, configuration, and seed; selected episode identities; client-call counts and times by role; graph diagnostics; and JSON episode outcomes. Client meters exclude provider-internal retries and token counts. The analysis script matches dataset, split, scene, episode, and seed, rejects omitted episodes, and resamples paired scene clusters. With fewer than two scenes it returns no interval. The experiments here use observed ranges and point differences; bootstrap intervals are not included.

\subsection{Runtime}
\begin{table}[t]
\centering
\caption{Component runtime and memory usage at different episode lengths. VLM timing is per call and does not include the full navigation pipeline.}
\label{tab:scalability}
\begin{tabular}{lrrr}
\toprule
Episode length & 10 & 20 & 40\\
\midrule
Memory (MB) & 1.2 & 2.8 & 5.1\\
Graph update (ms) & 12 & 18 & 25\\
VLM inference (ms) & 2300 & 2300 & 2300\\
\bottomrule
\end{tabular}
\end{table}
These measurements do not establish end-to-end latency, asymptotic scaling, or total evaluation cost.

\section{Additional Analyses}

\subsection{Roles of Coarse and Fine Stages}

The stages are intended to rank directions and check compatibility with the prior. The current component comparison does not separately measure coarse-stage recall, fine-stage precision, pruning error rates, or loop frequency. It therefore cannot explain the results through a proven division of labor. In particular, prior consistency is not a ground-truth test: correct but unusual observations can be rejected, and plausible errors can remain.

\subsection{Full Component Configuration}
Table~\ref{tab:ablation_details} restates the three memory configurations. Basic VLM uses no persistent memory; the voxel configuration retains direct VLM scoring; HGWM changes both memory and decision scoring.

\begin{table}[t]
\centering
\caption{Memory configurations on HM3D v0.2~\citep{ramakrishnan2021habitat}. The final row changes both memory and scoring; it does not isolate their effects or measure ambiguity robustness.}
\label{tab:ablation_details}
\small
\begin{tabular}{lrr}
\toprule
\textbf{Configuration} & \textbf{SR(\%)$\uparrow$} & \textbf{SPL(\%)$\uparrow$} \\
\midrule
Basic VLM / no memory & 67.8 & 29.3 \\
VLM / voxel memory & 70.9 & 30.1 \\
\textbf{HGWM / graph matching} & \textbf{74.4} & \textbf{31.9} \\
\bottomrule
\end{tabular}

\end{table}

\subsection{Additional Component Results}
Table~\ref{tab:additional_components} compares the component ablations with full HGWM at 74.4\% SR and 31.9\% SPL. The non-adaptive weight setting is evaluated as a single control. The voxel-memory baseline uses direct VLM scoring and obtains 70.9\% SR and 30.1\% SPL.

\begin{table}[t]
\centering
\caption{Additional HM3D v0.2 component results.}
\label{tab:additional_components}
\begin{tabular}{p{0.55\linewidth}rr}
\toprule
Configuration & SR (\%) & SPL (\%)\\
\midrule
Full HGWM & 74.4 & 31.9\\
Without goal prior & 72.1 & 30.8\\
Without relation refinement & 72.1 & 30.4\\
Without LLM arbitration & 72.7 & 31.1\\
With non-adaptive fine weights & 73.5 & 31.3\\
\bottomrule
\end{tabular}
\end{table}

Removing the prior decreases SR/SPL by 2.3/1.1 points; removing relation refinement decreases them by 2.3/1.5 points. Removing arbitration and using non-adaptive weights yield decreases of 1.7/0.8 and 0.9/0.6 points, respectively. These comparisons favor the complete configuration, although the effects of individual removals need not be additive.

\subsection{Prior Information and Graph Matching}
The goal prior has two roles: text conditioning for the coarse VLM and a relational reference for symbolic scoring. Prompting can reason about these relations implicitly, while the matching stage converts their compatibility with observations into an explicit directional score. The two channels therefore use the same prior information through different computations.

\paragraph{Comparison with prompting.}
Matching the scene and goal graphs improves SR by 1.5 percentage points and SPL by 1.4 points over prior prompting. Table~\ref{tab:matching_attribution} summarizes this gain alongside the memory and refinement comparisons. The matching result is expressed as a difference rather than split-specific absolute scores. A stricter controlled comparison would help separate the direct scoring effect from related changes in planning.

Such a comparison would use identical episode sets, model versions, sampling settings, and cached goal priors. Memory construction, relation refinement, access to graph context, map updates, stopping rules, and the local action interface would remain unchanged.

A common phase rule independent of fine scores, together with an identical arbitration prompt that omits those scores, would remove two indirect paths through which matching affects decisions. The two configurations would then differ in whether fusion uses the fine score or the equal-coefficient coarse substitution defined in the main paper. Clipping and the coarse-score override would remain unchanged. Because actions can diverge, subsequent observations need not be identical. This design defines a further experiment rather than a reinterpretation of the current full-model result.

\paragraph{Scope of the control settings.}
The existing prior-prompt-only switch also removes persistent graph context, refinement, and arbitration. The graph-memory/direct switch preserves graph context but removes refinement, fine scoring, and arbitration. Neither is the matched comparison just defined. The no-prior and no-refinement rows answer narrower intervention questions; they cannot substitute for a direct test of explicit matching. This stricter protocol is a proposed extension to the existing control settings.

\begin{table}[t]
\centering
\caption{Memory, refinement, and matching comparisons. Gains and drops are percentage-point differences. The matching row summarizes an additional comparison and is not an absolute SR/SPL pair.}
\label{tab:matching_attribution}
\begin{tabular}{p{0.64\linewidth}l}
\toprule
Quantity & Evidence\\
\midrule
Full vs. voxel-memory SR/SPL gains & 3.5/1.8 pp\\
Removal of refinement SR/SPL drops & 2.3/1.5 pp\\
Scene--goal matching vs. prompting: SR/SPL gains & +1.5/+1.4 pp\\
\bottomrule
\end{tabular}
\end{table}

\subsection{Three-Run Range on HM3D v0.1}
We evaluate the main Gemini-1.5-Pro configuration three times on HM3D v0.1. Relative to the reference scores of 62.6 SR and 31.5 SPL, both metrics vary by $-0.3$ to $+0.2$ percentage points. The observed ranges are therefore 62.3--62.8 SR and 31.2--31.7 SPL. These marginal minima and maxima may occur in different runs and are not confidence intervals or standard deviations. The range applies only to this configuration. Its SPL lower bound equals the WMNav point estimate, motivating a cautious interpretation of the small SPL gain.

\subsection{Further Evaluation}
Replaying recorded states with the same prior, graph, coarse scores, phase, and LLM output would help characterize the direct numerical matching channel. Replacing $10bf^{\alpha}$ with $bc^{\alpha}$ would reveal changes in ranking before and after clipping and map updates. This diagnostic would complement navigation experiments, since fixed-state replay cannot estimate closed-loop SR or SPL.

Further comparisons could examine graph and voxel memory under the same decision rule, coarse-only and fine-only scoring, and joint component removals. Matching episodes, prompts, seeds, and inference budgets would improve comparability. Annotations of relation updates, revisitation rates, and failure categories would also help explain when the method succeeds and where it fails.

\paragraph{Robustness beyond the current setting.}
Controlled depth and localization perturbations, unusual room--object arrangements, and dynamic scenes are useful next steps beyond the present simulation experiments. Physical-robot evaluation would additionally test the effects of sensing, communication, and motion latency on the complete system.

\section{Limitations and Future Work}

\textbf{Current Limitations.}
\begin{itemize}
    \item \textit{Simulation-only Evidence:} No physical-robot results or controlled noise robustness tests are available. Backbone sensitivity is evaluated for only two VLM settings on one split.
    \item \textit{Component Attribution:} The memory baselines change representation and scoring together. Additional component results are available, but their interactions and statistical uncertainty remain unestablished.
    \item \textit{Static Environments:} Our current framework assumes static scenes. Dynamic objects (e.g., people, pets) or movable furniture could violate the structural priors encoded in $G_{goal}$.
    \item \textit{Foundation Model Dependency:} Performance is constrained by the capabilities of the underlying VLMs/LLMs. Detection errors can propagate to graph construction; heuristic pruning can also remove correct atypical observations.
    \item \textit{Computational Cost:} Despite caching optimizations, real-time deployment remains challenging due to VLM inference latency (~2.3s per call).
\end{itemize}

Future studies could evaluate temporal graph updates and search for multiple targets. Active perception and physical robot trials are also untested directions.

\begin{figure*}[t]
\centering
\includegraphics[width=0.31\textwidth]{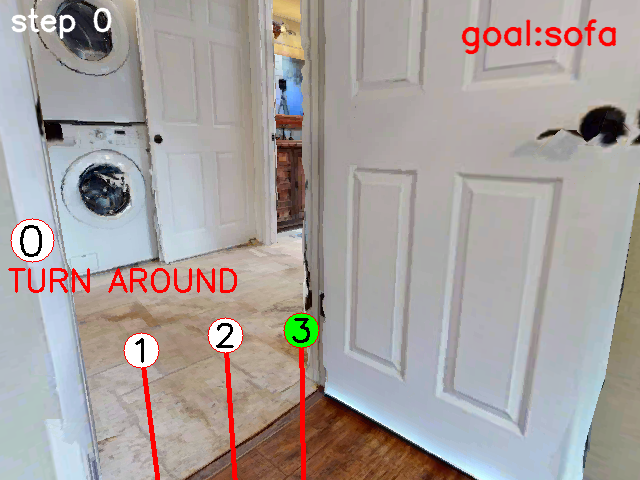}\hfill
\includegraphics[width=0.31\textwidth]{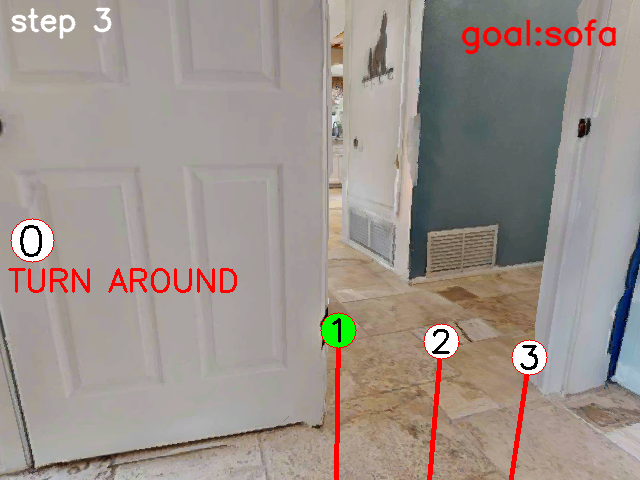}\hfill
\includegraphics[width=0.31\textwidth]{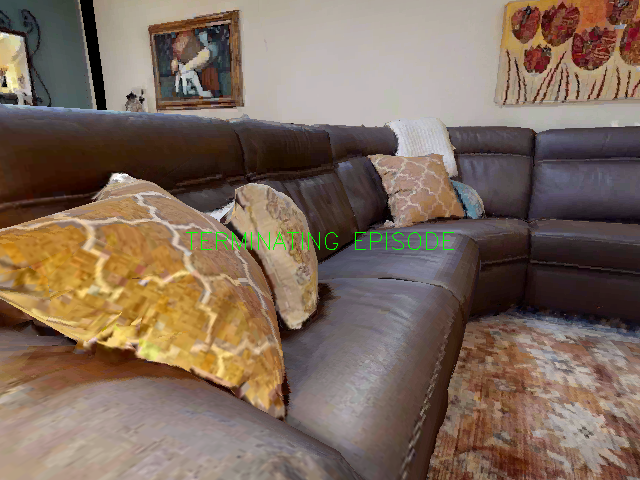}\\
\includegraphics[width=0.31\textwidth]{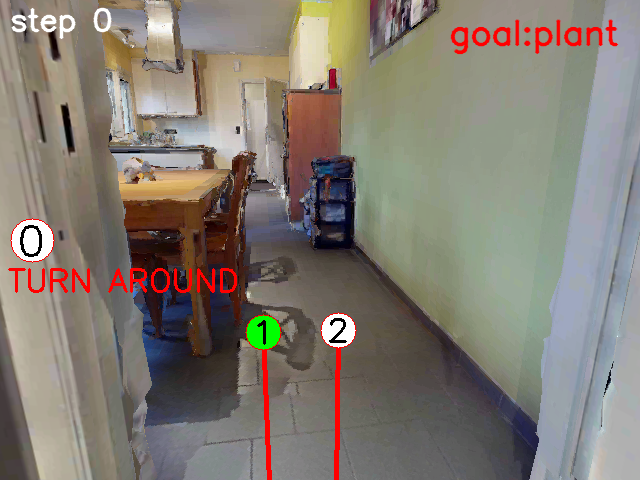}\hfill
\includegraphics[width=0.31\textwidth]{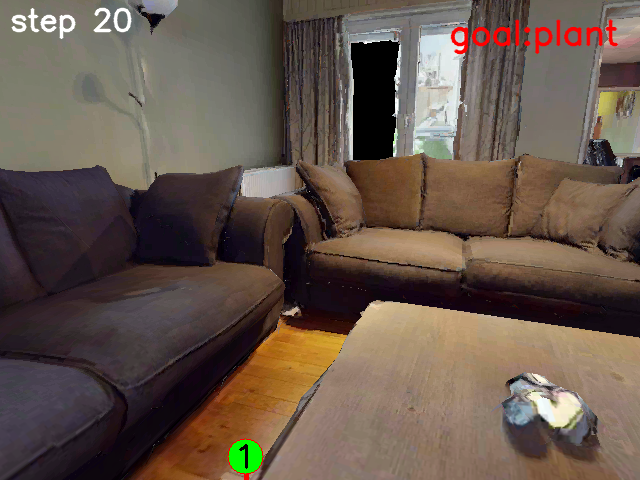}\hfill
\includegraphics[width=0.31\textwidth]{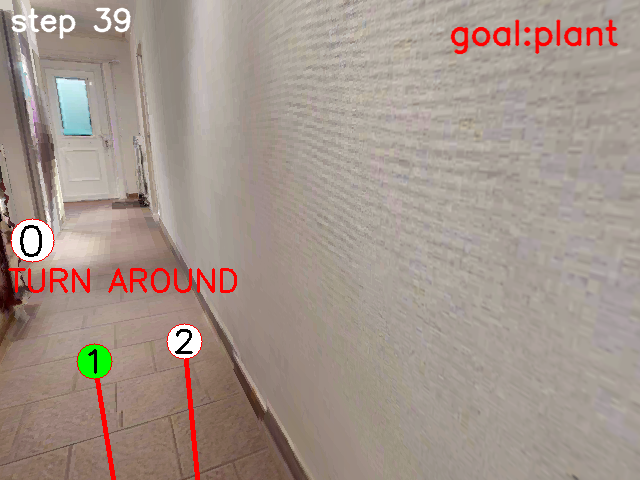}
\caption{Unedited archived views. Top: case 115\_853 at steps 0, 3, and 5. Bottom: case 220\_890 at steps 0, 20, and 39. Text and action annotations are part of the original images. The examples illustrate how observations evolve during a short and a longer search.}
\Description{Six original navigation frames from two archived cases, with their original target and action annotations.}
\end{figure*}

\section{Implementation Availability}
The implementation at \url{https://github.com/itak04/HGWM} includes configuration files, prompt builders, ablation settings, and analysis scripts. Experiment records store source and configuration identifiers, episode identities, and model-call statistics. The code and case collection provide the implementation details and qualitative context for the experiments. Access to benchmark datasets and model services follows the respective providers' terms.

\paragraph{LLM roles and writing assistance.}
Model roles and executable prompts are documented in the code. AI-assisted tools were used for language editing and formatting. The authors take responsibility for the methodology, results, and conclusions.

\paragraph{Code availability.}
The code directory contains a README, implementation notes, source index, and configuration files. The project page links the manuscript, source, and case gallery.

\section{Archived Navigation Cases}
\label{app:cases}
The supplementary archive covers 18 navigation cases. It indexes 338 step directories and retains 337 available text logs. Table~\ref{tab:case_inventory} lists the coverage, including one missing text record. Open the case gallery linked from \url{https://itak04.github.io/HGWM/} to browse each case and its prompts, responses, selected views, curiosity maps, and navigation maps.

\begin{table}[t]
\centering
\caption{Navigation case collection. Counts refer to recorded steps and available text logs.}
\label{tab:case_inventory}
\begin{tabular}{llrr}
\toprule
Case & Logged target & Steps & Logs\\
\midrule
115\_853 & sofa & 6 & 6\\
160\_853 & tv screen & 7 & 7\\
199\_890 & plant & 29 & 29\\
220\_890 & plant & 40 & 40\\
249\_890 & tv screen & 40 & 40\\
265\_890 & toilet & 19 & 19\\
289\_890 & chair & 40 & 40\\
297\_843 & toilet & 8 & 8\\
2\_877 & bed & 4 & 4\\
310\_843 & sofa & 2 & 2\\
32\_877 & chair & 6 & 6\\
352\_843 & toilet & 15 & 15\\
382\_843 & bed & 24 & 24\\
399\_824 & bed & 14 & 14\\
450\_824 & bed & 40 & 40\\
546\_880 & chair & 4 & 4\\
594\_829 & chair & 10 & 9\\
63\_877 & sofa & 30 & 30\\
\bottomrule
\end{tabular}
\end{table}

\paragraph{Selection and fidelity.}
The original collection is about 913 MB. To fit the supplementary limit, the first, middle, and last available step of each folder are selected by one deterministic rule, without consulting outcomes. The selected PNG files are copied byte-for-byte. All available step text is preserved, with a manifest recording source checksums and inclusion decisions. Additional frames and GIFs are omitted; pickle files are not deserialized or required by the gallery. The original source directory is unchanged.

\paragraph{What these records establish.}
The logs expose what the model was prompted with and what it returned. For example, case 115\_853 begins with an open-door subtask and ends with a logged sofa locator response; the final image visibly contains a sofa. Such observations help inspect a decision sequence, but a model goal flag or termination overlay is not a verified simulator success value. Some logged prior summaries contain zero node/edge counts; these summaries are retained as written and are not assumed to fully describe a hidden runtime graph.

The cases provide qualitative examples rather than a benchmark sample for estimating SR, SPL, or failure frequencies. We retain error-marked steps and identify missing text records in the index to make the selection process transparent.

\clearpage
\bibliographystyle{plainnat}
\bibliography{hgwm}
\end{document}